\documentclass[dvipsnames,format=sigconf,anonymous=false,review=false]{acmart}

\AtBeginDocument{%
  \providecommand\BibTeX{{%
    \normalfont B\kern-0.5em{\scshape i\kern-0.25em b}\kern-0.8em\TeX}}}

\copyrightyear{2023}
\acmYear{2023}
\setcopyright{rightsretained}
\acmConference[GECCO '23 Companion]{Genetic and Evolutionary Computation
Conference Companion}{July 15--19, 2023}{Lisbon, Portugal}
\acmBooktitle{Genetic and Evolutionary Computation Conference Companion
(GECCO '23 Companion), July 15--19, 2023, Lisbon,
Portugal}\acmDOI{10.1145/3583133.3590731}
\acmISBN{979-8-4007-0120-7/23/07}

\begin{document}

\title[Generative Adversarial Neuroevolution for Control Behaviour Imitation]{Generative Adversarial Neuroevolution \\for Control Behaviour Imitation}
\author{Maximilien Le Clei}
\affiliation{%
  \institution{Montreal University Geriatric Institute}
  \city{Montréal}
  \country{Canada}
}
\email{maximilien.le.clei@umontreal.ca}

\author{Pierre Bellec}
\affiliation{%
  \institution{Montreal University Geriatric Institute}
  \city{Montréal}
  \country{Canada}
  \postcode{H3W 1W5}
}
\email{pierre.bellec@criugm.qc.ca}
\renewcommand{\shortauthors}{Le Clei and Bellec}

\begin{abstract}
There is a recent surge in interest for imitation learning, with large human video-game and robotic manipulation datasets being used to train agents on very complex tasks. While deep neuroevolution has recently been shown to match the performance of gradient-based techniques on various reinforcement learning problems, the application of deep neuroevolution techniques to imitation learning remains relatively unexplored. In this work, we propose to explore whether deep neuroevolution can be used for behaviour imitation on popular simulation environments. We introduce a simple co-evolutionary adversarial generation framework, and evaluate its capabilities by evolving standard deep recurrent networks to imitate state-of-the-art pre-trained agents on 8 OpenAI Gym state-based control tasks. Across all tasks, we find the final elite actor agents capable of achieving scores as high as those obtained by the pre-trained agents, all the while closely following their score trajectories. Our results suggest that neuroevolution could be a valuable addition to deep learning techniques to produce accurate emulation of behavioural agents. We believe that the generality and simplicity of our approach opens avenues for imitating increasingly complex behaviours in increasingly complex settings, e.g. human behaviour in real-world settings. We provide our source code, model checkpoints and results at \textcolor{blue}{\href{https://github.com/MaximilienLC/gane/}{github.com/MaximilienLC/gane/}}.
\end{abstract}

\maketitle

\section{Introduction}

In order to generate strong behavioural agents, deep neuroevolution works often target reinforcement learning problems. Various evolutionary algorithms have recently been shown to be capable of optimizing deep neural network parameters, at times matching the performance of gradient-based techniques on popular modern reinforcement learning benchmarks \cite{salimans2017evolution,such2017deep,risi2019deep, leclei2022neuroevolution}. Yet, deep neuroevolution techniques are at times unable to produce strong agents across full sets of reinforcement learning benchmarks, even when evolving large populations over thousands of generations \cite{salimans2017evolution,such2017deep, leclei2022neuroevolution}. This phenomenon is not unique to neuroevolution: deep learning algorithms are also known to struggle with reinforcement learning problems in which reward signals are sparse and provide limited information about the current state of optimization \cite{lecun2016predictive}. 

In this work, we propose to explore whether neuroevolution is capable of producing strong behavioural agents through self-supervised imitation learning rather than reinforcement learning. Specifically, we introduce a generative adversarial co-evolutionary optimization framework wherein two equal sized populations of agents repeatedly match against each other over the course of the evolutionary process. The first population consists of agents generating data points that can take any format (text, audio, behaviour, etc), which we call generator agents, while the second population contains agents, which we call discriminator agents, tasked to discriminate between generated data points and data points sampled from some target distribution. In order to evaluate the capabilities of this optimization framework, we propose to imitate the behaviours of state-of-the-art pre-trained deep reinforcement learning agents on various popular state-based control tasks available through the OpenAI Gym API \cite{brockman2016openai}. Across all tasks, as the adversarial optimization process unfolds, we find generator agents to obtain increasingly high scores, with the final elite agents matching pre-trained agent performance across all tasks while closely following their target agents’ score trajectories.

\vspace{-1em}

\section{Background}

Reinforcement learning \cite{sutton2018reinforcement} is an artificial intelligence paradigm in which artificial agents learn to maximize some notion of cumulative reward in an environment. These quantitative signals are often hand-crafted to proxy for qualitative estimations of what constitutes valuable behaviours. And while much success has been found through this learning paradigm over the past couple of years \cite{mnih2013playing,berner2019dota,vinyals2019grandmaster}, it is known to possess two major shortcomings. First, it has been observed that optimizing for reward signals is prone to produce both unexpected and undesired behaviour \cite{clark2016faulty}. Secondly, in many reinforcement learning problems, reward signals are sparse and provide limited amounts of valuable information for optimization algorithms, especially when contrasted with supervised and self-supervised learning paradigms \cite{lecun2016predictive}.

While reinforcement learning is often the preferred choice to generate strong behavioural agents through deep neuroevolution, other learning paradigms could be of interest. Imitation learning, for instance, is a powerful and practical alternative in which agents are optimized to produce sequential decision-making policies given some existing target behaviour \cite{hussein2017imitation}. Bypassing the need for a handcrafted reward function and the challenges it carries, this learning paradigm has recently been applied to various complex problems \cite{torabi2019recent}, such as video-game playing \cite{baker2022video} and robotic manipulation \cite{brohan2022rt}.  

However, many imitation learning techniques fall under the supervised learning regime and therefore require handcrafting a similarity metric between behaviours, potentially also leading to unexpected and undesired behaviour. To address this shortcoming, recent research has shown that self-supervised learning techniques like generative adversarial networks can be utilized instead, and have been shown to be capable of generating strong control task behaviours across diverse tasks \cite{ho2016generative}. 

While self-supervised imitation learning has, to the best of our knowledge, not been explored in the context of neuroevolution, several works have already harnessed self-supervised learning techniques for tasks like image generation \cite{costa2020neuroevolution}. In the broader evolutionary computation context, such multi-population evolutionary techniques pertain to the field of competitive co-evolution which has in the past been explored both in relation to dynamically complexifying neural networks \cite{stanley2004competitive} and deep neural networks for high-dimensional problems \cite{klijn2021coevolutionary}. Finally, it was also recently argued \cite{arulkumaran2019alphastar} that competitive co-evolution techniques were actually an integral to recent state-of-the-art game-playing agent AlphaStar \cite{vinyals2019grandmaster}.

\vspace{-1em}
\section{Evolutionary Adversarial Generation}

\textbf{Overview.} We now present a generic co-evolutionary framework for generative adversarial optimization, that purposely neither imposes restrictions on the agent inner mechanisms, nor the data format, nor the proceedings of both variation and selection evolutionary stages. We maintain, in this evolutionary framework, two separate populations of agents. The first population consists of generator agents tasked to generate data points, which can in principle fall under any modality: image, text, behaviour, etc. The second population is composed of discriminator agents tasked to discriminate between generated data points and data points originating from some target distribution.

\textbf{Agent interactions.} For every iteration of the evolutionary process, during the agent evaluation stage, each discriminator is randomly matched to one unique generator and one data point drawn from the target distribution $p_T$. In a given match, generator $G$ first produces data point $x_G$. In turn, discriminator $D$ observes both $x_G$ and a true target point $x_T$. For each datapoint, the discriminator outputs a confidence estimate that the point truly originated from the target distribution $p_T$ rather than the generator $G$, resulting in two scores: $D(x_G)$ and $D(x_T)$. This confidence estimate is then used to compute the fitnesses of both generator and discriminator agents, as follows.

\textbf{Agent fitnesses.} Making use of discriminator outputs $D(x_G)$ and $D(x_T)$, the generator’s fitness $fit_G$ is first set to $D(x_G)$, meaning that it is proportionally rewarded by the discriminator’s incorrect assessment of $x_G$. In turn, the discriminator’s fitness $fit_D$ is contrasting both prediction scores, $D(x_T) - D(x_G)$, meaning that it is increasingly rewarded for correctly assessing $x_T$ and increasingly penalized for incorrectly assessing $x_G$.

We have outlined here a version of the co-evolutionary framework wherein generators and discriminators are only paired once. However it is quite straightforward to increase the number of pairings between agents to construct fitnesses from a larger number of interactions, thus trading off execution speed with lower bias.

\section{Experiments}

\vspace{-1em}
\begin{figure}[H]
  \centering
  \includegraphics[width=1\linewidth]{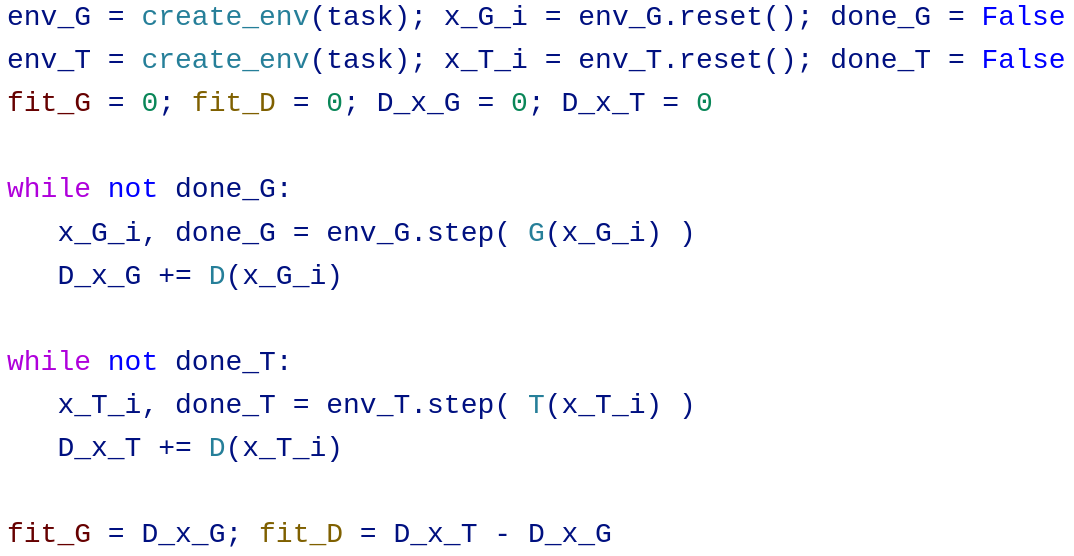}
  \vspace{-2em}
  \caption{Agent evaluation pseudo-code. \normalfont Trimmed code snippet of the generator \& discriminator evaluation for a given task.}
\end{figure}

\vspace{-1em}

\textbf{Task environments.} In order to evaluate the adversarial generation framework described above, we propose to run imitation learning experiments on a subset of the state-based control task environments available in OpenAI Gym \cite{brockman2016openai}. We select 8 state-based tasks with a pre-trained agent available (see section below) and input/output dimensions amenable to rapid experimentation. Across these environments, agents get to control simplified virtual robotic arms, vehicles and simulated bodies. In order to perform these control tasks, agents are iteratively fed input values characterizing various pieces of information like the position, angle and speed of their body parts. In turn, agents are expected to output values representing various actuation forces onto their body components. These environments also provide termination criteria and reward signals as a means to drive behaviour in reinforcement learning settings. Finally, environment instances can be slightly altered (agent starting position, etc) to assess out of distribution generalisation.

\textbf{Target and Evolved Agents.} We retrieve pre-trained agents from the “RL Baselines3 Zoo'' library \cite{rl-zoo}. The 8 agents (1 per task) achieve state-of-the-art performance on their respective tasks and are optimized through various popular deep reinforcement learning algorithms \cite{mnih2013playing,schulman2017proximal,haarnoja2018soft,fujimoto2018addressing,kuznetsov2020controlling}. In order to imitate these agents, we employ the co-evolutionary adversarial framework to optimize the parameters of static deep recurrent neural networks. We set up 2 populations, each composed of 64 neural network agents. As all tasks make use of different input and output sizes, we equip all generator agents with recurrent networks of dimension (d\_input, 50, 50, d\_output) and all discriminator agents with recurrent networks of dimension (d\_input, 50, 50, 1),  the last hidden layer being recurrent for both types of agents, similar to the networks described by Salimans et al. \cite{salimans2017evolution} in their experiments. As is common practice \cite{goodfellow2016deep}, recurrent layers make use of hyperbolic tangent activations while dense layers make use of rectified linear units (ReLU). 

\textbf{Evolved agent inputs and outputs.} As input distributions vary between environments, we implement across all control tasks a running standardization of inputs for all generator and discriminator agents. In environments requiring continuous action values, we clip the ReLU activated values emitted from the generators’ network’s output nodes in the range [0, 1] and scale them to the expected range of outputs. In environments requiring discrete action values, we instead feed the index of the output node emitting the largest value. In turn, the discriminators' network’s ReLU activated outputs are clipped in [0,1]. In order to emit prediction scores, the discriminator agents’ outputs are averaged over the total number of episode states experienced by the observed (generator or pre-trained) agent to emit a final prediction score also in range [0,1].

\textbf{Evolutionary algorithm.} We now detail how the co-evolutionary framework was implemented for the 8 control tasks. As a proof of feasibility, we demonstrated the adversarial learning framework using  a basic genetic algorithm, stripped of crossover and speciation mechanisms, iterating over variation, evaluation and selection stages as described in \cite{such2017deep}. All neural network weights and biases are initialized at 0 and perturbed every generation, during the variation stage, by values sampled from $\mathcal{N}(0,0.01)$, where $\mathcal{N}(\mu,\sigma^{2})$ is the normal distribution with mean $\mu$ and variance $\sigma^{2}$. During the evaluation stage, on each task, we set both the generator and pre-trained agents to produce behaviour in their own instances of the same task environment. Finally, during the selection stage, we make use of a 50\% truncation selection within each agent population, meaning that the top 50\% agents in terms of fitness score are selected and duplicated over the bottom 50\% agents.

\vspace{-2pt}
\section{Results}
\vspace{-1em}

\begin{figure}[H]
\centering
\includegraphics[width=\linewidth]{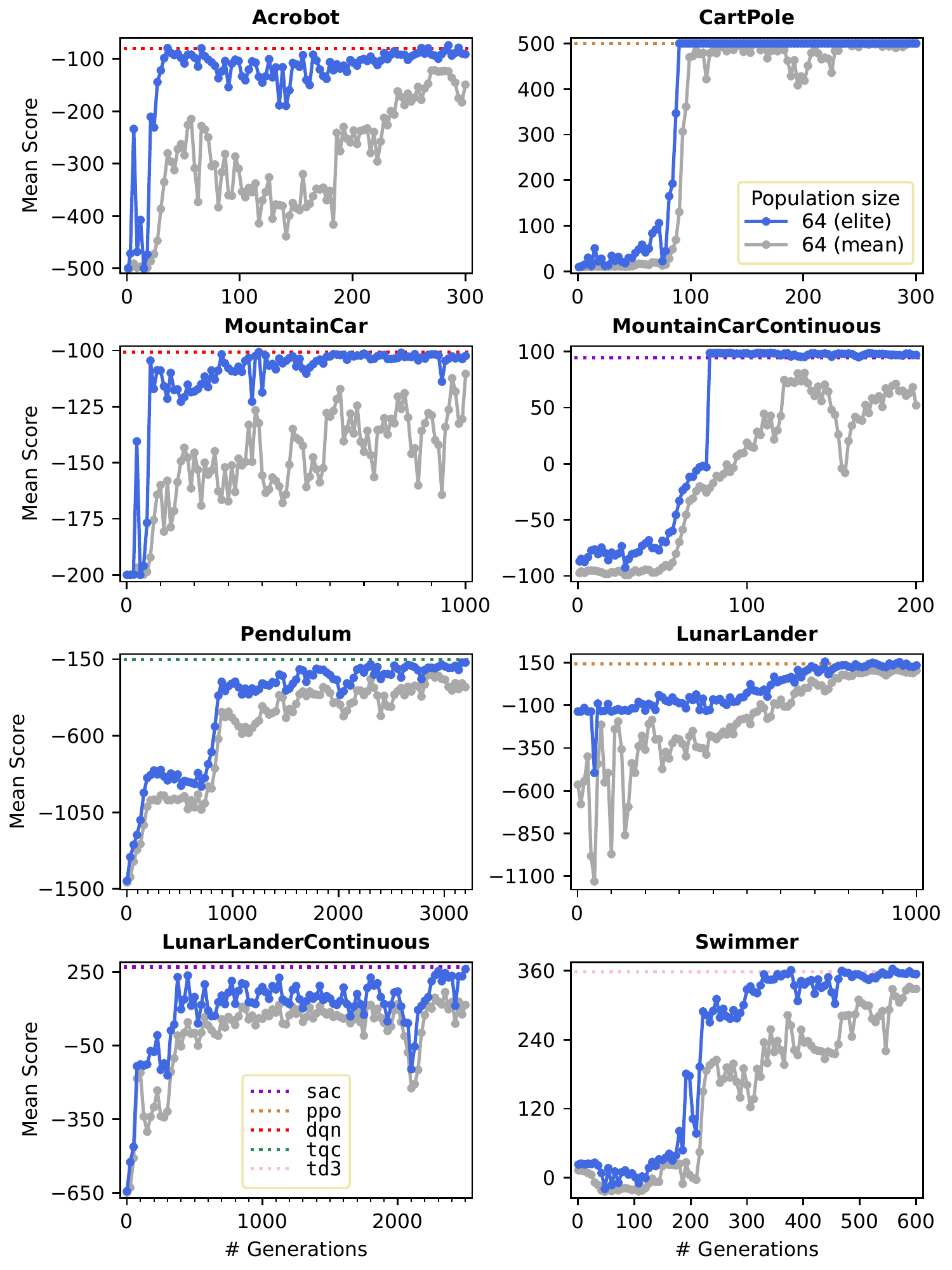}
\vspace{-2em}
\caption{Averaged scores. \normalfont Progression of the mean population and elite scores, averaged over 10 environment instances never observed during optimization. Across all tasks, elite generator agents match the target pre-trained agents’ performances. The population generally does not lag far behind.}
\label{result1.pdf}
\end{figure}
\vspace{-1em}
\textbf{Evolution of mean scores.} Our main aim is to assess whether agents evolved through the adversarial framework are able to generate behaviour comparable to that of their targets. Figure \ref{result1.pdf} shows the progress of both the whole population and its elite agent, averaged over 10 environment instances never observed during the evolutionary process. On all 8 tasks, we find the generator agents capable of achieving as high scores as the target pre-trained deep reinforcement learning baselines. In certain environments, such as \textit{Acrobot}, \textit{CartPole} and \textit{MountainCarContinuous}, the process is quite fast, with elite (highest performing) generator agents reaching pre-trained agent performance in less than 100 generations. However, in other environments, such as \textit{LunarLanderContinous} and \textit{Pendulum}, this process is quite slower, taking up to a few thousand generations for elite generator agents to match target performance. We also observe that in certain tasks like \textit{Acrobot} and \textit{LunarLanderContinuous}, performance at times plunges before rising again, suggesting that in certain periods of the evolutionary process, generator agents were required to experiment with initially detrimental behaviour shifts in order to fool their now more competent paired discriminator agents. Finally, the generator populations consistently follow quite closely the elite scores across all 8 tasks, suggesting that elite generator agents are not simple outliers and that the entire generator populations are indeed evolving towards high scores.

\vspace{-1em}

\begin{figure}[H]
\centering
\includegraphics[width=\linewidth]{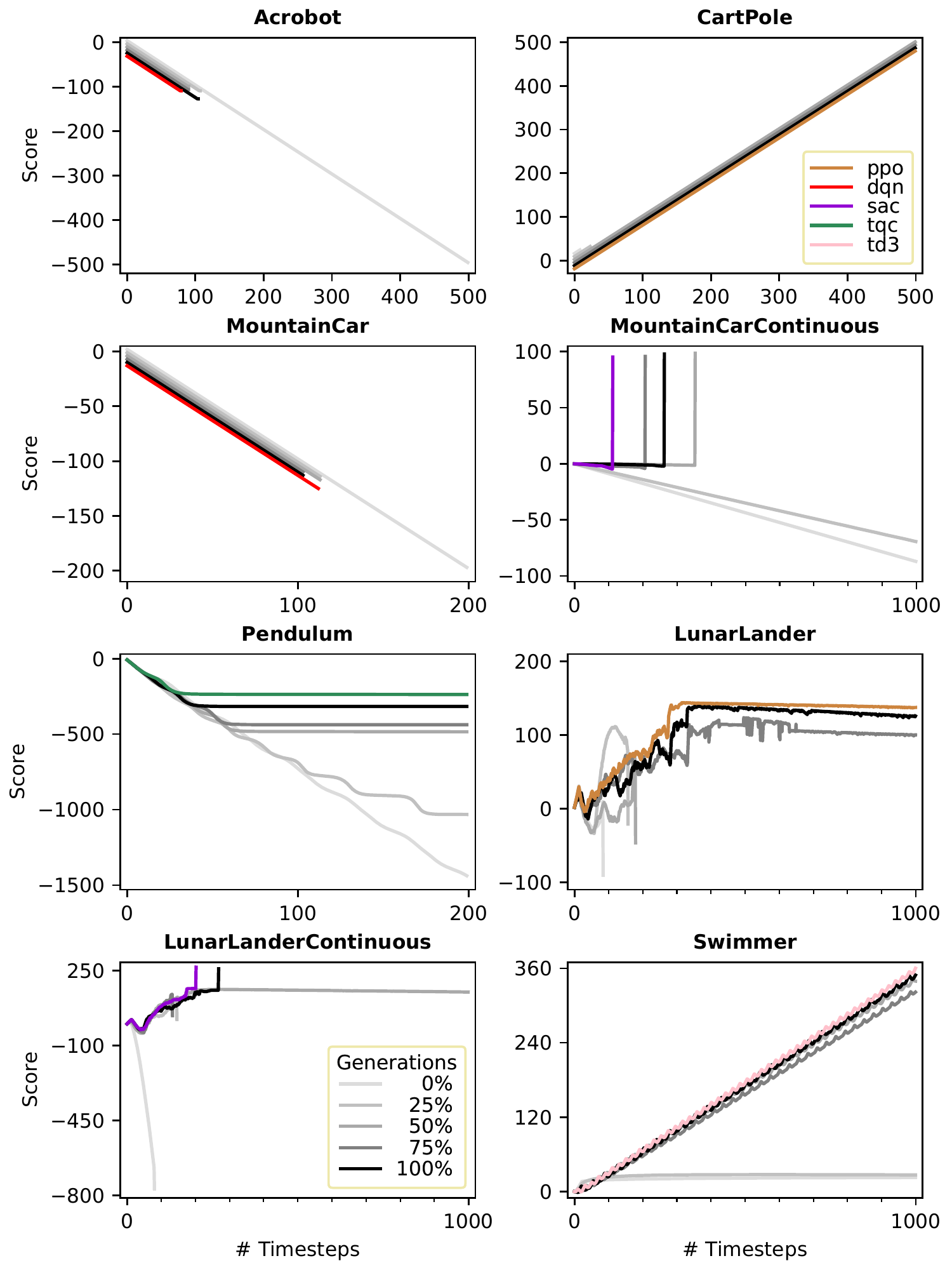}
\vspace{-2em}
\caption{Elite agent score trajectories. \normalfont Elite generator agents tend to increasingly closely match the score trajectories of their target. The score trajectories for Acrobot, MountainCar, CartPole have been slightly shifted to prevent them from overlapping.}
\label{result2.pdf}
\end{figure}

\textbf{Score trajectories.} There often exists a wide range of policies capable of achieving high scores in reinforcement learning environments. To understand the heterogeneity of policies in our agent population, Figure \ref{result2.pdf} shows the progression of agent scores over the course of one episode, running each task on an environment instance never observed during the optimization process. For each task, we show the progression of the target pre-trained agent as well as elite generator agents at various stages of the evolution process, namely the 1st generation, and 25\%, 50\%, 75\% and 100\% of the total number of generations. We systematically observe that, as the number of generations increases, score trajectories of evolved agents become increasingly similar to those of the target pre-trained agents. In \textit{Acrobot} and \textit{MountainCar}, constant negative rewards are emitted until termination, through either success or time expiration. Elite agents are able to terminate the episode at an increasingly similar timestep to that of the pre-trained agent. In \textit{CartPole}, the reward function emits constant positive rewards until termination by failure or time expiration. Elite agents increasingly hold off failure until time expiration. In \textit{MountainCarContinuous}, rewards are emitted until termination by either success or time expiration. Elite agents are able to complete the task at an increasingly similar pace to that of the pre-trained agent. In task \textit{LunarLanderContinuous}, the reward function emits rewards until termination by success, failure or time expiration. Elite agents go from failing, to reaching time expiration, to finally succeeding and closely following the score trajectory of the pre-trained agent. Finally, in \textit{Pendulum} and \textit{Swimmer}, the reward function emits rewards until time expiration. Elite agents' score trajectories also become increasingly similar to that of the pre-trained agent.

\section{Discussion}

We have observed, throughout our experiments, generated behaviours increasingly similar to their targets according to both their final score and score trajectories. These results seem to indicate that the generative adversarial evolution framework is capable of evolving generator agents able to emulate various types of behaviour, all the while utilizing very limited information signals. Indeed, the generator agents are in particular blind to both their own score and the particular behaviors of the target agent they are trying to imitate. The generator agent population instead evolves in complete reliance on prediction scores emitted by discriminator agents from successive single generator-discriminator matches. This condensed and potentially noisy information however appears sufficient to produce accurate behaviors across these control tasks. 

We also remark that this framework is quite flexible and thus could be used in many other evolutionary settings, and we believe that many improvements over the basic implementation described here are possible. For instance, calculating agent fitnesses from more than a single random generator-discriminator pairing could stabilize the evolutionary process. Last but not least, making use of dynamic network architectures \cite{leclei2022neuroevolution} instead of static network architectures could potentially bring a multitude of additional benefits such as higher compression and faster inference.

Within a broader machine learning context, we believe our work to bring out a different approach to the imitation learning problem. Indeed, deep imitation learning is often framed as a supervised learning problem wherein networks are trained to model, to various degrees, state-action pair probabilities. Our framework instead approaches the imitation learning problem from a broader point of view, wherein agents are instead evolved to model full episodes of behaviour, and therefore hypothesize that our approach could enable the imitation of higher-level characteristics of behaviour.

\section{Conclusion}

We introduced in this work a simple yet general evolutionary adversarial generation framework. We evaluated its capabilities by evolving deep recurrent network agents to imitate full episodes of pre-trained reinforcement learning agents on various state-based control tasks. We found the evolved generators to produce similarly strong behaviour across all the tasks and increasingly follow the pre-trained agents' score trajectories. We believe our work opens avenues to imitate more valuable behaviours, such as human or animal behaviours in order to solve currently challenging tasks.

\renewcommand*{\bibfont}{\scriptsize}
\bibliographystyle{ACM-Reference-Format}
\bibliography{refs}

\end{document}